\documentclass[12pt]{l4dc2021}

\makeatletter
\def\set@curr@file#1{\def\@curr@file{#1}} 
\makeatother
\usepackage[load-configurations=version-1]{siunitx} 

\usepackage{jmlrutils} 
\usepackage{algorithm}
\usepackage{algorithmic}
\usepackage{xcolor}
\usepackage{hyperref}
\usepackage{caption}

\title{Generating Adversarial Disturbances  for Controller Verification} 
\usepackage{times}

 \author{\Name{Udaya Ghai}$^{1}*$ \Email{ughai@cs.princeton.edu}\\
  \Name{David Snyder}$^{2}*$ \Email{dasnyder@princeton.edu}\\
  \Name{Anirudha Majumdar}$^{2,3}$ \Email{ani.majumdar@princeton.edu}\\
  \Name{Elad Hazan}$^{1,3}$ \Email{ehazan@cs.princeton.edu}\\
  $*$ \addr Equal contribution \\
  \addr $1$ Department of Computer Science, Princeton University \\
  $2$  Department of Mechanical and Aerospace Engineering, Princeton University \\
  $3$ Google AI Princeton
  }

\usepackage{amsfonts}
\usepackage{mathtools}
\usepackage{latexsym}
\usepackage{relsize}
\usepackage{sidecap}

\usepackage[capitalise]{cleveref}
\usepackage{xcolor}
\usepackage{dsfont}
\usepackage{multicol}

\newcommand{\treg}{\text{TrustRegion}}
\newcommand{\reshape}{\text{reshape}}

\def\hess{\nabla^2}
\def\grad{\nabla}

\newcommand{\A}{\mathcal{A}}

\def\regret{\mbox{{Regret}}}

\newcommand{\ignore}[1]{}

\newcommand{\email}[1]{\texttt{#1}}

\newtheorem{assumption}[theorem]{Assumption}

\DeclareMathAlphabet{\mathbfsf}{\encodingdefault}{\sfdefault}{bx}{n}

\DeclareMathOperator*{\vect}{vec}
\DeclareMathOperator*{\Exp}{Exp}

\DeclareMathOperator{\RR}{\mathbb{R}}

\newcommand{\lrnorm}[1]{\left\|#1\right\|}

\newcommand{\ceil}[1]{\lceil #1 \rceil}

\newcommand{\wt}[1]{\smash{\widetilde{#1}}}

\renewcommand{\O}{O}

\newcommand{\tO}{\wt{\O}}

\newcommand{\E}{\mathbb{E}}

\newcommand{\poly}{\mathrm{poly}}

\newcommand{\eps}{\varepsilon}

\newcommand{\eqdef}{:=}

\renewcommand{\leq}{~\le~}
\renewcommand{\geq}{~\ge~}

\let\oldtfrac\tfrac
\renewcommand{\tfrac}[2]{\smash{\oldtfrac{#1}{#2}}}

\let\nablaold\nabla
\renewcommand{\nabla}{\nablaold\mkern-2.5mu}

\newcommand{\pa}[1]{\left(#1\right)}

\newcommand{\inp}[2]{\left\langle #1,#2\right\rangle}

\newcommand{\mZ}{\mathcal{Z}}

\newcommand{\mO}{\mathcal{O}}

\newcommand{\mL}{\mathcal{L}}

\newcommand{\citeFullPaper}{An extended version of this paper is available online \citep{Ghai20} and contains proofs and implementation details. References to the Appendix correspond to this extended version.}

\renewcommand{\citeFullPaper}{}

\begin{document}
\maketitle
\thispagestyle{empty}


\begin{abstract}
We consider the problem of generating maximally adversarial disturbances for a given controller assuming only blackbox access to it. We propose an online learning approach to this problem that \emph{adaptively} generates disturbances based on control inputs chosen by the controller. The goal of the disturbance generator is to minimize \emph{regret} versus a benchmark disturbance-generating policy class, i.e., to maximize the cost incurred by the controller as well as possible compared to the best possible disturbance generator \emph{in hindsight} (chosen from a benchmark policy class). In the setting where the dynamics are linear and the costs are quadratic, we formulate our problem as an online trust region (OTR) problem with memory and present a new online learning algorithm (\emph{MOTR}) for this problem. We prove that this method competes with the best disturbance generator in hindsight (chosen from a rich class of benchmark policies that includes linear-dynamical disturbance generating policies). 
We demonstrate our approach on two simulated examples: (i) synthetically generated linear systems, and (ii) generating wind disturbances for the popular PX4 controller in the AirSim simulator.
On these examples, we demonstrate that our approach outperforms several baseline approaches, including $H_{\infty}$ disturbance generation and gradient-based methods. 

\end{abstract}


\section{Introduction}

We consider the problem of certifying the safety and correct operation of control algorithms in the context of robotics, as understood by a measure of the worst-case system performance in the presence of uncertainty and disturbances. Motivated by this challenge, we consider the following idealized problem. 

Consider a control system given by $x_{t+1} = f(x_t, u_t, w_t)$, with state $x \in \mathcal{X} \subseteq \mathbb{R}^{d_x}$, control input $u \in \mathbb{R}^{d_u}$, and disturbance $w \in \mathbb{R}^{d_w}$.  Suppose we are provided \emph{blackbox access} to a controller for this system, i.e., we do not have access to the software that defines the controller, but can observe the closed-loop system's behavior by choosing disturbance values. The controller may be arbitrarily complex (e.g., adaptive, nonlinear, stateful, etc.). Our goal is to generate disturbances $w_t$ that \emph{maximize} a specified cost $\sum_{t=0}^\infty c(x_t, u_t)$ incurred by the controller. 

\paragraph{Statement of Contributions.} We present an online learning approach for tackling the problem of generating disturbances for dynamical systems in order to maximize the cost incurred by a controller  assuming only blackbox access to it. The key idea behind our approach is to leverage techniques from {online learning} (see e.g. \cite{hazan2019introduction}) to \emph{adaptively} choose disturbances based on control inputs chosen by the controller. Determining the optimal disturbance for a given controller with online blackbox access is computationally infeasible in general. We thus consider \emph{regret minimization} versus a benchmark disturbance-generating policy class. 
Since our goal is to maximize the cost of the controller, the natural formulation of our problem in online learning is non-convex. Online non-convex optimization does not admit efficient algorithms in general. 

To overcome this challenge, we consider the case when the system is linear and the costs are quadratic. In this case we formulate our problem as a special case of non-convex optimization, namely an online trust region (OTR) problem with memory.  
We then present a new \emph{online trust region with memory} algorithm (\emph{MOTR}) with optimal regret guarantees, which may be of independent interest. 
Using this technique, we prove that our method competes with the best disturbance-generating policy {in hindsight} from a reference class. This reference class includes all state-feedback linear-dynamical policies (Def.~\ref{LDDG:def}).

\begin{figure}[t]
    \centering
    \includegraphics[width=0.7\linewidth]{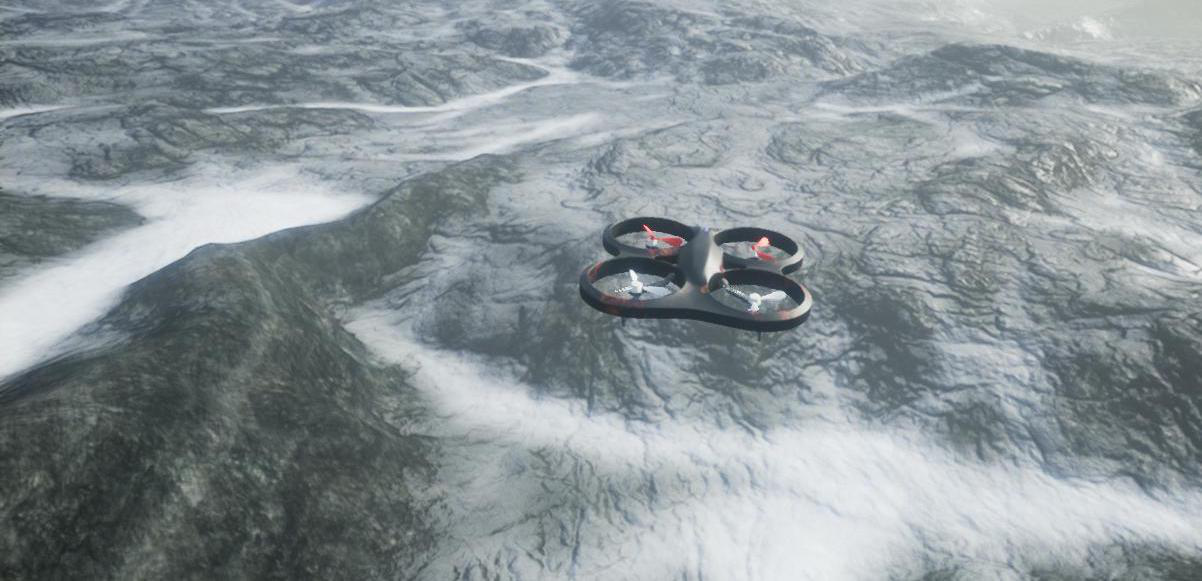}
        \caption{\footnotesize{Quadrotor in AirSim Mountains Environment. We consider the problem of generating adversarial disturbances (e.g., wind gusts) for a given controller.}}
    \label{fig:AirSimAnchor}
\end{figure}

We demonstrate our approach on two simulated examples: (i) synthetically generated linear systems, and (ii) generating wind disturbances for the highly-popular PX4 controller \citep{px4} in the physically-realistic AirSim drone simulator \citep{airsim} (Fig.~\ref{fig:AirSimAnchor}). 
We compare our approach to several baseline methods,  including gradient-based methods and an  $H_{\infty}$ disturbance generator. For linear systems, $H_{\infty}$ is a Nash equilibrium solution to the offline disturbance problem; however, this does not hold for the case of time-varying costs or when the controller deviates from an $H_\infty$ paradigm. We demonstrate the ability of our method to adaptively generate disturbances that outperform these baselines. 

\citeFullPaper

\subsection{Related Work}

\paragraph{Regret minimization for online control.}
There is a large body of work within the control theory literature on synthesizing robust and adaptive controllers (see, e.g., \cite{Stengel1994OptimalCA,kemin}). The most relevant work for our purposes is online control with low \emph{regret}. 
In classical control theory, the disturbances are assumed to be i.i.d. Gaussian and the cost functions are known ahead of time. 
In the \emph{online} LQR setting \citep{abbasi2011regret,dean2018regret,mania2019certainty,cohen2018online}, a fully-observed linear dynamic system is driven by i.i.d. Gaussian noise and the learner incurs a quadratic state and input cost. 

Recent algorithms \citep{mania2019certainty,cohen19b,cohen2018online} attain $\sqrt{T}$ regret
for this online setting, and are able to cope with changing loss functions.
\cite{agarwal2019online} consider the  more general and challenging setting of
\emph{non-stochastic control} in which the disturbances are adversarially
chosen, and the cost functions are arbitrary convex costs. The key insight behind this result is
an improper controller parameterization, known as disturbance-action control, coupled with
advances in online convex optimization with memory due to \cite{anava2015online}.
Non-stochastic control was extended to the setting of unknown systems and partial observability
\citep{hazan2019nonstochastic,simchowitz2020improper}.

In contrast to the work mentioned above, we consider the problem of generating adaptive \emph{disturbances} that \emph{maximize} the cumulative cost incurred by a given controller. This shift in problem formulation introduces fundamental technical challenges. In particular, the primary challenge is the \emph{non-convexity} associated with the cost maximization problem. Providing regret guarantees (from the point of view of the disturbance generator) in this non-convex setting constitutes one of they key technical contributions of this work. 

\paragraph{Adversarial reinforcement learning.} The literature on generating disturbances for
control systems is sparse compared to the body of work on synthesizing robust controllers. There
has been recent work on generating adversarial policies for agents trained using reinforcement
learning (motivated by a long line of work on generating adversarial examples for supervised
learning models \citep{goodfellow_15}). These results consistently suggest that for
high-dimensional problems in RL settings, non-adversarially-trained agents can be directly
harmed in training \citep{behzadan_17} and are highly susceptible to multiple adversarial
failure modes \citep{huang_17,gleave_20}. The latter problem motivates
\cite{vinitsky_Robust_20} to train agents against an ensemble of adversaries to generate a more
robust learned policy. In this vein, \cite{mandlekar_17} integrates the ideas into a robust training algorithm that allows for noise perturbations in the standard control formulation. However, in contrast to our work, none of the above robust training protocols make theoretical guarantees about the performance of their trained agent, whereas we are able to obtain explicit regret guarantees for the performance of our adversarial agent. 

A parallel line of investigation uses sampling-based techniques for probabilistic safety assurance \citep{sinha_Neural_20} by actively seeking samples of rare events to estimate the probability of failure modes. This is extended in \cite{sinha_FZ_20} to include online methods. In particular, they generate models of multiple adversaries (akin to \cite{vinitsky_Robust_20}) and then use online learning to `decompose' their observed adversary into elements of their set of modeled adversaries, choosing robust actions accordingly. This differs from our work in the optimization paradigm. In particular, they require Monte Carlo sampling of multiple trajectories, repeated over subsequent updates of the environment distribution parameters, in order to obtain guarantees. Our guarantees are `within-trajectory,' in that we learn and compete with a class of disturbance generators within a single trajectory, rather than by optimizing over many simulations. 
\paragraph{Online learning and the trust region problem.}
We make extensive use of techniques from the field of online learning and regret minimization in games \citep{cesa2006prediction,hazan2016introduction}. Most relevant to our work is the literature on online non-convex optimization \citep{agarwal2019learning,suggala2019online}, and online convex optimization with memory \citep{anava2015online}. 
The problem of maximizing a general quadratic function subject to Euclidean norm constraint is known as the Trust Region (TR) problem, which originated in applying Newton's method to non-convex optimization. Despite the non-convex objective, TR is known to be solvable in polynomial time via a semi-definite relaxation \citep{Aharon_1996}, and also allows for accelerated gradient methods \citep{Hazan_2015}.

\vspace{-3mm}
\section{Setting and Background}
\vspace{-1mm}
\subsection{Notation}
For vectors, we use the notation $\|x\|_Q = \sqrt{x^{\top} Q x}$ for a weighted euclidean 
norm, where $Q$ is a positive definite matrix. For matrices, we use $\|M\|_F$ to denote 
the Frobenius norm of matrix $M$ and $\|M\|$ to denote the spectral norm.
\vspace{-3mm}
\subsection{Setting}
We consider a nonstochastic 
 linear time-invariant (LTI) system defined by the following equation:
 \vspace{-1mm}
\begin{equation*}
x_{t+1} = A x_t + B u_t + Cw_t,~ 
\end{equation*}
\vspace{-1mm}
where $x_t \in \RR^{d_x}$ is the state, $u_t \in \RR^{d_u}$ is the control 
input, and $w_t \in \RR^{d_w}$ is the adversarially chosen disturbance.  We assume the state and disturbance dimensions are 
the same and $C = I$ for the 
remainder of this exposition, but the results still hold in full 
generality. At time $t$, a quadratic cost $c_t(\cdot, \cdot)$ 
is revealed and the controller suffers cost $c_t(x_t, u_t)$. 
As the disturbances are adversarially chosen, the trajectory, and thus the costs 
are determined by this. In this model, a disturbance generator $\A$ is a (possibly randomized) 
mapping from all previous states and actions to a disturbance vector. As the goal is to 
produce worst-case disturbances, the cost $c_t$ for the controller is a reward for $\A$. 
The states produced by $\A$ with controls $u_t$ are denoted $x^{\A}_t$ and the 
total reward is denoted
\vspace{-1mm}
\begin{equation*}
J_T(\A) = \sum_{t=1}^T c_t(x^{\A}_t, u_t) = \sum_{t=1}^T \|x^{\A}_t\|^2_{Q_t} + \|u_t\|^2_{R_t}
~.
\end{equation*}
For a randomized generator, we consider the expected reward. We use a \emph{regret} 
notion of performance, where the goal is to play a disturbance sequence $\{w_t\}_{t=1}^T$ 
such that the reward is competitive with reward corresponding to the disturbances 
played by the best fixed disturbance generator $\pi$, chosen in hindsight from a comparator class $\Pi$.
\vspace{-1mm}
\begin{equation*}
\regret_T(\A) = \max_{\pi \in \Pi} J_T(\pi) - J_T(\A).
\end{equation*}
\vspace{-1cm}
\subsection{Comparator class}
For our comparator class, we consider a bounded set of Control-disturbance Generators, 
defined as follows.
\begin{definition}\label{CDG:def}
A Control-disturbance Generator (CDG), $\pi(M)$ is specified by parameters 
$M = (M^{[1]}, \dots ,M^{[H]})$, along with a 
bias\footnote{The bias term is not included in the remainder of the 
theoretical work for simplicity, but equivalent results can be proved including bias.} 
$w_0$, where the disturbance $w_t$ played at state $x_t$ is defined as
\begin{align}
  w_t = \sum_{i=1}^H M^{[i]} u_{t-i} + w_0~.
\end{align}
\end{definition}

\begin{definition}\label{comparator:def}
Let $\Pi_{D, H} = \{\pi(M): M \in \RR^{Hd_x \times d_u}, \|M\|_F \leq D\}$ 
be the set of CDGs with history $H$ and size $D$. We also use the shorthand $M \in \Pi_{D,H}$.
\end{definition}

A CDG is the equivalent of a Disturbance-action controller (DAC) \citep{agarwal2019online} 
where the roles of actions and disturbances have been swapped. It has been shown that
DACs can approximate Linear Dynamic Controllers (LDCs), a powerful generalization of 
linear controllers. As such, we analogously define Linear Dynamic Disturbance Generators (LDDGs), 
which likewise are approximated by CDGs.

\begin{definition}[Linear Dynamic Disturbance Generator]
\label{LDDG:def}
A \emph{linear dynamic disturbance generator} $\pi$ is a linear dynamical system 
$(A_{\pi}, B_{\pi}, C_{\pi}, D_{\pi})$ with internal state $s_t \in \RR^{d_{\pi}}$, input $x_t \in \RR^{d_{x}}$, and output $w_t \in \RR^{d_w}$ that satisfies
\vspace{-3mm}
\begin{align*}
  s_{t+1} = A_{\pi} s_t + B_{\pi} x_t,~ w_{t} = C_{\pi} s_t + D_{\pi} x_t \ .
\end{align*}
\end{definition}
\vspace{-0.8cm}
\subsection{Assumptions}
We make the following assumptions requiring an agent to play bounded controls and requiring bounded system and cost matrices:
\begin{assumption}[Bounded controls]\label{bounded-controls:as}
The control sequence is bounded so $\|u_t\| \leq C_u$.
\end{assumption}
\vspace{-0.4cm}
\begin{assumption}[Stabilizable Dynamics\label{stab_dy:as}]
Consider the dynamics tuple $\{A, B, C, u(x,t)\}$. We assume that $A = HLH^{-1}$, with $ \|L\| \leq 1 - \gamma$,  matrix $A$ having condition number $\|H\|\|H^{-1}\|\leq \kappa$ and $\|A\|, \|B\|, \|C\| \leq \beta$.
\end{assumption}
Note that if $A$ is not open-loop stable, but the pair $(A, B)$ is stabilizable, there exists a matrix $K^*$ such that $\tilde{A} = A-BK^*$ satisfies the above criterion, and we can equivalently analyze the system $\{\tilde{A}, B, u^*(x,t)\}$, where $u^*(x,t) = u(x,t)+K^*x$. This transformation explains the generality of Assumption \ref{stab_dy:as}. Importantly, we do not require knowledge of $u(x,t)$ in this transformation. 

\vspace{-1mm}
\begin{assumption}[Bounded costs]\label{bounded_costs:as}
The cost matrices have bounded spectral norm, $\|Q_t\|,\|R_t\| \leq \xi$.
\end{assumption}
\vspace{-3mm}
\ignore{
Following \cite{Chen2020BlackBoxCF}, we use $\mL$ to denote the complexity of the system and comparator class where
\begin{align*}
  \mL = d_x + d_u + D + \beta + \kappa + \xi + \gamma^{-1}, \text{where}
\end{align*}
\begin{itemize}
  \item $d_x =$ dimension of states $x_t\in \RR^{d_x}$.
  \item $d_u =$ dimension of actions $u_t\in \RR^{d_u}$.
  \item $C_u =$ upper bound on magnitude of controls.
  \item $C_x =$ upper bound on magnitude of states.
  \item $D =$ the bound on the comparator.
  \item $H = \ceil{\gamma^{-1}\log(\kappa \xi T)}$ the history of the CDG comparator set.
  \item $\Pi = \Pi_{D,H}$ is the comparator disturbance policy set.
  \item $\beta =$ upper bound on spectral norm of system matrices.
  \item $\gamma =$ lower bound on $ 1- \|L\|$.
  \item $\kappa =$ condition number of $A$.
  \item $\xi =$ upper bound on spectral norm of costs.
\end{itemize}
}

Following \cite{Chen2020BlackBoxCF}, we use $\mL$ to denote the complexity of the system and comparator class where 
$\mL = d_x + d_u + d_w + D +C_u + \beta + \kappa + \xi + \gamma^{-1}$. 
Here, $d_x, d_u, d_w$ are the state, action, and disturbance dimensions
respectively. $C_u$ bounds the magnitude of controls. The comparator class is $\Pi_{H,D}$ with $H =
\ceil{\gamma^{-1}\log(\kappa \xi T)}$, in order to capture LDDGs. $\beta$ and
$\xi$ are spectral norm bounds on system matrices and costs respectively. The
condition number and decay of dynamics are $\kappa$ and $\gamma$ respectively.

\section{Online Trust Region With Memory}\label{online_trust_region:sec}
This section describes our main building block for the adversarial disturbance generator: 
an online non-convex learning problem called \emph{online trust region (OTR) with
memory}. In Sec.~\ref{tr_background:sec} we provide background on the trust
region problem. Subsequently, in
Sec.~\ref{otrm:sec} we formally introduce the online trust region with
memory setting. 

\subsection{Trust Region Problem}\label{tr_background:sec}
We show that the cost of a CDG can be approximated closely by a nonconvex quadratic. Optimizing the cost of the policy is then a trust region problem,  a well-studied quadratic optimization over a Euclidean ball. The interest in this problem stems from the fact that it is one of the most basic non-convex optimization problems that admits ``hidden-convexity" --- a property that allows efficient algorithms that converge to a global solution. 

\begin{definition}
A trust region problem is defined by a tuple $(P,p,D)$ with $P \in \RR^{d \times d}$, $p \in \RR^d$, and $D > 0$ 
as the following mathematical optimization problem
\begin{align*}
    \max_{\|z\| \leq D} \quad  z^{\top} P z + p^{\top} z .
\end{align*}
\end{definition}

We can define a condition number for a trust region problem as follows.
\begin{definition}
The condition number for a trust region instance $(P,p, D)$ is $\kappa = \frac{\lambda}{\mu}$, where
$\lambda = \max(2(\|P\|_2 + \|p\|_2), D, 1)$ and  $\mu =  \min(D, 1)$.
\end{definition}

Note that a trust region problem can be solved in polynomial time by conversion to an equivalent convex optimization problem (see e.g. \cite{Aharon_1996}). 

\begin{theorem}\label{tregion_opt:theorem}
Let $(P,p,D)$ be a trust region problem with condition number $\kappa$. There exists
an algorithm $\treg$ such that $\treg(P,p, D, \epsilon)$ produces $z_{a}$ with $\|z_{a}\| \leq D$ such that
\begin{align*}
    z_{a}^{\top}Pz_{a} + p^{\top}z_{a} \geq \max_{\|z\| \leq D} z^{\top}Pz + p^{\top}z - \epsilon ~,
\end{align*} 
and runs in time $\poly(d, \log \kappa, \log \frac{1}{\eps})$.
\end{theorem}

\subsection{Online trust region with memory}\label{otrm:sec}
Consider the setting of online learning, where an algorithm $\A$
predicts a point $z_t$ with $\|z_t\| \leq D$. We use the shorthand, 
$z_{t:H} = (z_{t-H+1}, \dots ,z_t) \in \RR^{dH}$ for the concatenation of the $H$ last
points. The algorithm then receives feedback from an adversarially chosen quadratic reward
function $f_t: \RR^{dH} \rightarrow \RR$ of the last $H$ decisions, parameterized by 
$P_t \in \RR^{dH \times dH}$ and $p_t \in \RR^{dH}$. The reward 
function is defined as 
\begin{align}\label{quad_loss:eq}
    f_t(z')= {z'}^{\top} P_t z' + p_t^{\top}z'.
\end{align}

The reward function acting on a single point is also useful, so we define $g_t: \RR^{d} \rightarrow \RR$ with
\begin{align}\label{trust_reg_g:eq}
    g_t(z) &= f_t(z,\dots z)
    = (z,\dots z)^{\top} P_t (z,\dots z) + p_t^{\top} (z,\dots z)
    \eqdef z^{\top} C_t z + d_t^{\top}z.
    \end{align}

The reward earned in round $t$ is  $f_t(z_{t:H})$. The goal of the online player is to  minimize the expected regret, compared to playing the single best point in hindsight:
\begin{align}\label{trust_region_regret:eq}
    \regret(\A) =  \max_{\|z\| \leq D}\sum_{t=H}^T g_t(z) - \E[\sum_{t=H}^T f_t(z_{t:H})]~.
\end{align} 
Here the expectation is over randomness of the algorithm. In App.~\ref{otrm_alg:sec}, we provide a polynomial-time $O(H^{\frac{3}{2}}T^{\frac{1}{2}})$ regret algorithm for the online trust region with memory. Below is an informal statement of this result (See Thm.~\ref{otrm:theorem} for the full result).

\begin{theorem}\label{otrm_informal:theorem}
Suppose elements of matrices $P_t$ and elements of $p_t$ bounded.
Alg.~\ref{OTR:alg}, with suitable parameterization, will incur expected regret at most
\begin{align*}
    \regret(T) \leq O(D^2 d^{5/2}H^{3/2}\sqrt{T})~,
\end{align*}
and the runtime of the algorithm for each iteration will be $\poly(d,H,\log D, \log T)$.
\end{theorem}
The algorithm works by applying an extension of nonconvex Follow-the-Perturbed-Leader (FPL) \citep{agarwal2019learning,suggala2019online} to functions with memory (see App.~\ref{fplm:sec}). In the OTR with memory setting, the perturbed subproblems that need to be solved in each iteration are trust region problems, so they can be solved in polynomial time.

\vspace{-5pt}
\section{Algorithm and Main Theorem}\label{algo:sec}
\vspace{-2pt}
Our algorithm (\emph{MOTR}; Alg.~\ref{ODG:alg}) is an application of Alg.~\ref{OTR:alg} in the Appendix for the OTR with memory problem applied to approximations of the costs of playing a CDG, $\pi(M)$. Let $m = \vect(M)$ be a flattened version of the CDG matrix $M$. Our approximate cost, $g_t(m) = c_t(y_t(M), u_t)$ is the cost of an approximate state from a truncated rollout starting at $y_{t-H}(M) = 0$, with 
\begin{align}
    y_{s+1}(M) = Ay_s(M) + Bu_s + C\big ( \sum_{i=1}^H M^{[i]} u_{s-i} + w_0
    \big)~.
\end{align}

\begin{algorithm}[!htp]
\caption{Memory Online Trust Region (\emph{MOTR}) Generator}
\label{ODG:alg}
\begin{algorithmic}
\STATE \textbf{Input:} Rounds $T$, system parameters $(A,B, C)$,  noise parameter $\eta$, history $H$
\STATE Define $u_s =0$ for $s \leq 0$.
\STATE Initialize $M_0 \in \Pi_{D,H}$ randomly.
\STATE $S_{0} = 0_{Hd_xd_u,Hd_xd_u}, s_{0} =0_{Hd_xd_u}$.
\FOR {$t=0$ \TO $T$}
\STATE Observe quadratic reward $c_t$ and earn $c_t(x_{t}, u_t) = \|x_{t}\|^2_{Q_t} + \|u_t\|^2_{R_t}$
\STATE Generate disturbance $w_t = \sum_{i=1}^H M^{[i]}_t u_{t-i}$
\STATE Observe control $u_t$ and update state $x_{t+1} = Ax_t + Bu_{t} + w_{t}$
\STATE Define $g_t(m) = c_t(y_t(M), u_t)$ where
\begin{align*}
    y_{s+1}(M) = 
\begin{cases}
  Ay_s(M) + Bu_s + C\big ( \sum_{i=1}^H M^{[i]} u_{s-i} + w_0 \big) & s \geq t-h\\
  0 & \text{otherwise}
\end{cases}
\end{align*}
\STATE Define $S_t = S_{t-1} + (\hess g_t)(0)$ and $s_t = s_{t-1} + (\grad g_t)(0)$ [see \eqref{At:eq} and \eqref{bt:eq} in App.~\ref{main_proof:sec}]
\STATE Generate random vector $\sigma_t \in \RR^{Hd_xd_u}$ such that $\sigma_{t,i} \sim \Exp(\eta)$
\STATE Update $m_{t+1} \gets \treg(S_t, s_t - \sigma_t, D, \frac{1}{T})$
\STATE Reshape $M_{t+1} \gets \reshape(m_{t+1}, \RR^{Hd_x \times  d_u})$
\ENDFOR
\end{algorithmic}
\end{algorithm}

In App.~\ref{approx_state:sec}, we show that $g_t$ is a quadratic function of $m$, and that $y_t$ is an accurate approximation of $x_t$ due to the stabilizability of the dynamics. In App.~\ref{main_proof:sec}, we combine the regret bound for Alg.~\ref{OTR:alg} in  Thm.~\ref{otrm:theorem} with the approximation guarantee on $y_t$ from Lem.~\ref{approx_cost:lem}, yielding the following theorem.
\begin{theorem}
Suppose Assumptions~\ref{bounded-controls:as}, \ref{stab_dy:as}, \ref{bounded_costs:as} hold; then Alg.~\ref{ODG:alg} suffers regret at most $\tO(\poly(\mL)\sqrt{T})$.
\end{theorem}

\section{Experiments}
We evaluate the performance of the disturbance generator \emph{MOTR} defined in Alg. \ref{ODG:alg} across two settings. These consist of (1) general (randomly generated) linear systems of varying modal behavior, and (2) a thirteen-dimensional rigid-body model for a quadrotor drone in the AirSim simulation environment \citep{airsim}. 
To evaluate our algorithm in each setting, we compare its performance with the performance of several baseline disturbance generators, against several different controllers. In each setting, the \emph{MOTR} algorithm is able to outperform the baselines.

\subsection{Baseline Generators and Controllers}
We compare the \emph{MOTR} algorithm with five baseline generators. Sinusoidal and Gaussian noise are standard within control theory, and form the first two generator classes. A random directional  generator (a fixed-norm equivalent of the Gaussian generator) is third. The dynamic game formulation of the $H_\infty$ control problem \citep{basar_08, bernhard_HINF_91} yields a Nash equilibrium disturbance generator. The final baseline is a first-order online gradient ascent (OGA) policy, which does not provide theoretical guarantees in this nonconvex setting. OGA produces disturbances via a CDG, with the $M$ learned via gradient ascent on the instantaneous cost.

For the experimental settings in which the true dynamics are linear, the disturbance generators are tested across three controllers: a standard LQR controller, a $H_\infty$ infinite-horizon optimal controller, and an adaptive gradient perturbation controller (GPC) \citep{agarwal2019online}. In each setting, true, fixed system costs 
($\lrnorm{x}^2 + \lrnorm{u}^2$) were provided to the algorithms. 

In the AirSim experiment, the two nonlinear controllers tested are the Pixhawk PX4 controller \citep{px4}, which is one of the most popular controllers used by quadrotors in practice, and a pre-tuned PID controller that is defined by the AirSim environment.

\subsection{Notes on Implementation}
In order to ensure fair comparisons across the baselines, the actions chosen by each disturbance generator except the Gaussian are normalized to ensure that the available disturbance `budget' does not vary across generators. Thus, the sinusoid and random generators are essentially choosing directions within the state space. The Gaussian generator is specified so that its average disturbance norm will be slightly higher than the norm bound in expectation. The frequency and initial phase vector of each sinusoid generator are optimized offline against the open-loop system dynamics. Further details of the implementation of \emph{MOTR} are deferred to App.~\ref{appComp:section}

\subsection{Experiment 1: General Linear Systems}
Here, a randomly generated set of 11 linear systems are tested for each controller-generator pair over 10 initial conditions. For each system, $A \in \mathbb{R}^{4\times 4}$, $B,C \in \mathbb{R}^{4 \times 2}$. We define the cost of a trajectory to be equal to the cumulative average cost over the time horizon. For each controller-generator pair, we average the costs over the 10 initial conditions, and then normalize each generator's average cost for a given controller to lie in the range $[0, 1]$, where a higher value indicates stronger performance. These costs are aggregated across the 11 systems and scaled to the best-performing disturbance generator (for the given controller). The results are shown in Tab.~\ref{tab:4x2sys}.

\begin{table} [htbp!]
\centering
    \begin{tabular}{c||c|c|c|} 
      & LQR & GPC & $H_\infty$ \\ [0.5ex] 
     \hline \hline
     MOTR & $\mathbf{1.000 \pm 0.006}$ & $\mathbf{1.000 \pm 0.017}$ & 0.997 $\pm$ 0.038 \\ \hline
     OGA & 0.918 $\pm$ 0.128 & 0.897 $\pm$ 0.142 & 0.998 $\pm$ 0.038 \\ \hline
     $H_\infty$ & 0.980$\pm$ 0.035 & 0.949$\pm$ 0.107 & $\mathbf{1.000 \pm 0.039}$ \\ \hline
     Random & 0.328$\pm$ 0.097 & 0.323 $\pm$ 0.106 & 0.552 $\pm$ 0.112 \\ \hline
     Sine & 0.564 $\pm$ 0.297 & 0.540 $\pm$ 0.301 & 0.767 $\pm$ 0.264 \\ \hline
     Gaussian & 0.444 $\pm$ 0.160 & 0.438 $\pm$ 0.174 & 0.744 $\pm$ 0.196\\ \hline
     \end{tabular}
     \caption{\footnotesize{Performance results for disturbance generators aggregated over randomized linear systems. The time horizon is $T=200$, with 11 systems and 10 seeds per system.}\label{tab:4x2sys}}
     \vspace{-5mm}
\end{table} 
There are several important points to note. First, against an $H_\infty$ controller, the $H_\infty$ disturbance generator is a Nash equilibrium solution, so it is expected that \emph{MOTR} will recover but not exceed that performance. In addition to \emph{MOTR} strongly outperforming the Random, Sinusoid, and Gaussian generators in each setting, we see that against adaptive controllers like GPC, \emph{MOTR} also begins to outperform the $H_\infty$ disturbances. In the presence of model misspecification and cost mismatch, we expect this phenomenon may become more pronounced.

\vspace{-3mm}
\subsection{Experiment 2: Rigid-Body Drone in AirSim} 
Testing \emph{MOTR} within the AirSim environment allows an empirical test of several key elements of the algorithmic performance, including (1) scaling to higher system dimension, (2) generalizability to nonlinear dynamics about linearized reference conditions, and (3) performance with an accurate but low-dimensional model of the disturbance-to-state transfer function. 

The model follows the traditional rigid-body, 6 degree-of-freedom (6DOF) model for air vehicle dynamics, but uses quaternions instead of Euler angles, yielding a 13-dimensional state representation. The nonlinear dynamics are propagated about a nominal hover flight condition. This condition was linearized numerically using a least squares regression procedure on simulator data. There are four inputs, corresponding to the motor commands for each propeller of the quadcopter, and three disturbance channels, corresponding to the North-East-Down (NED) coordinates of the inertial wind vector. 
\vspace{-3mm}
\begin{figure}[!htbp]
\captionsetup[subfigure]{labelformat=empty}
    \centering
    \subfigure[] {
    \includegraphics[width=0.45\linewidth]{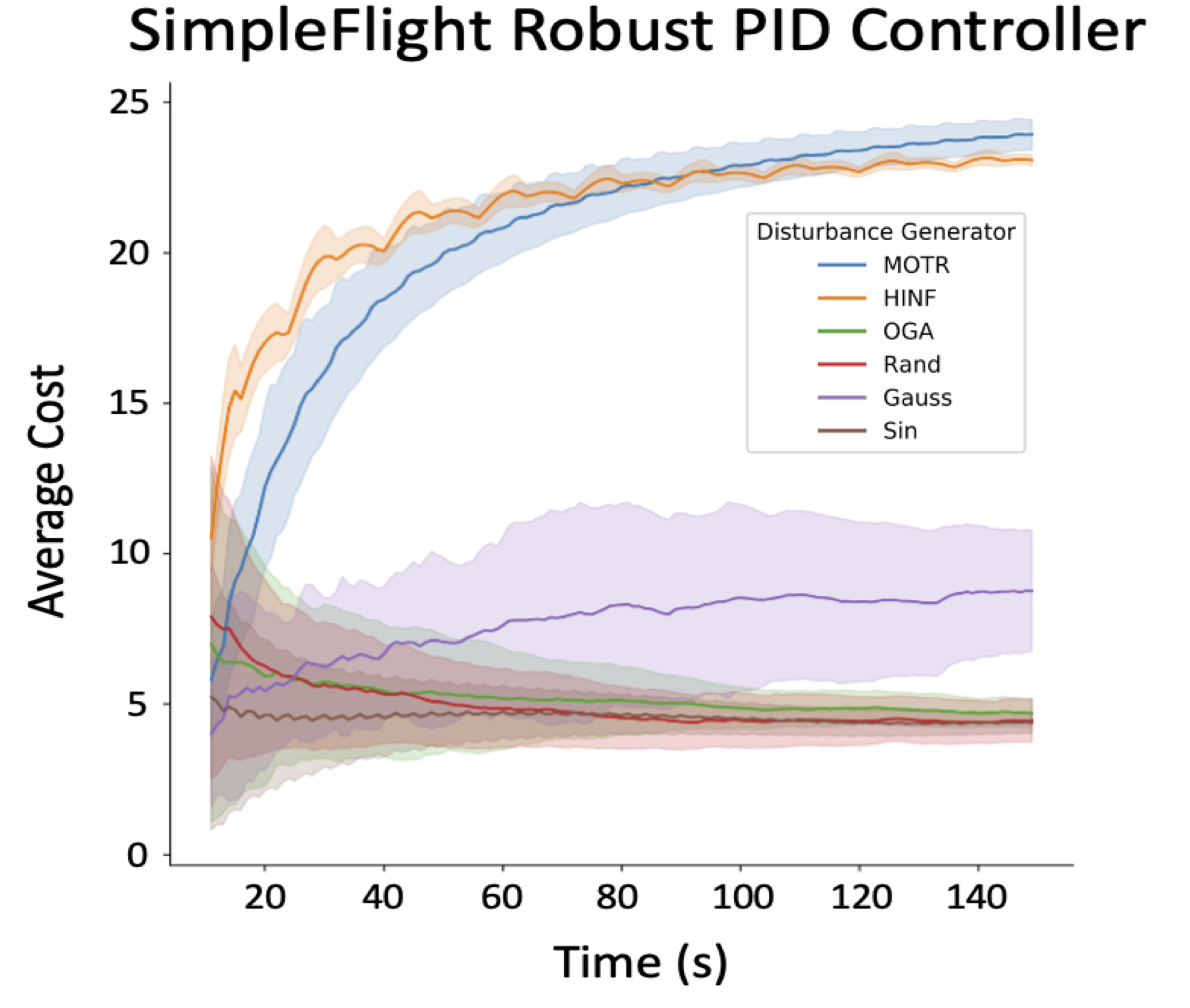}
    }
    \centering
    \subfigure[]{
    \includegraphics[width=0.42\linewidth]{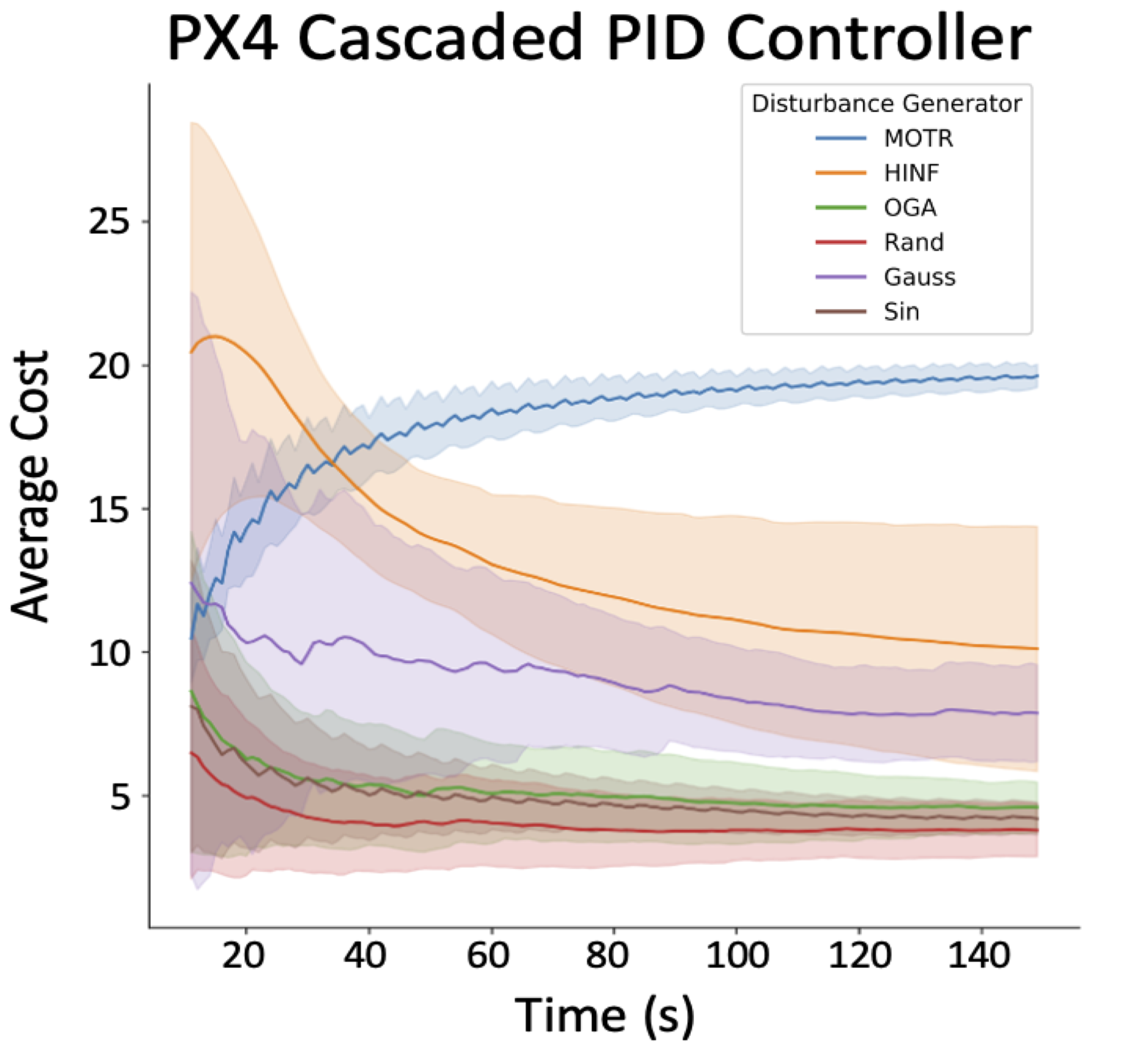}
    }
    \caption{\footnotesize{(a) Results for the SimpleFlight robust PID controller. Both $H_\infty$ and \emph{MOTR} perform well in this setting. Results are taken over 14 seeds per generator (b) Results for the popular PX4 controller. Note that the non-adaptive $H_\infty$ policy is attenuated, unlike \emph{MOTR}. Further, the first-order online method struggles in this setting. Results are taken over 15 seeds per generator.}}
    \label{airsim:fig}
  
\end{figure}

\newpage

The controllers utilized in the simulator include a `SimpleFlight' AirSim controller and the PX4 autopilot, which is incorporated into AirSim's software-in-the-loop PX4 stack. Each of these controllers is a nonlinear PID controller, and we note that the PX4 is one of the most commonly used autopilots for quadcopter drones. 

We present the results of the simulations in Fig.~\ref{airsim:fig}. An important feature of the simulations was the presence of a clear best strategy for large disturbances, corresponding to updrafts and downdrafts. However, because the wind magnitudes are constrained, the PX4 controller is able to adaptively attenuate this behavior. As such, the $H_\infty$ generator, while nearly as strong as \emph{MOTR} on the SimpleFlight simulations, suffered against PX4. \emph{MOTR}, however, was able to adapt in the PX4 setting and thus maintain strong disturbance performance. 
\section{Conclusions}

We have studied the problem of generating the worst possible disturbances for a given controller. 

This is a challenging non-convex problem, which we pose in the framework of online learning.
We describe a novel method based on regret minimization in non-convex games with provable guarantees. 
Our experimental results demonstrate the ability of our approach to outperform various baselines including gradient-based methods and an $H_{\infty}$ disturbance generator. 

This work raises many intriguing questions: can this approach be generalized to dynamics that are unknown, non-linear,  partially observable, admit bandit feedback and/or time-varying? Recent results in non-stochastic control suggest the feasibility of these directions \citep{Chen2020BlackBoxCF,gradu2020nonstochastic,gradu2020adaptive,simchowitz2020improper}. Another promising direction is to establish lower bounds on regret for the disturbance generation problem and find algorithms that match these lower bounds. Finally, in the vein of adversarial reinforcement learning, the inclusion of adversarial disturbances may prove a useful tool in synthesizing robust learned controllers, as having access to a an adaptive, online disturbance generation mechanism might enhance robustness and regularize worst-case behavior.


\clearpage

\acks{We thank Karan Singh for enlightening discussion. Elad Hazan is partially supported by NSF award \# 1704860 as well as the Google corporation. Anirudha Majumdar was partially supported by the Office of Naval Research [Award Number: N00014-18-
1-2873]. This material is based upon work supported by the National Science Foundation Graduate Research Fellowship Program under Grant No. DGE-2039656. Any opinions, findings, and conclusions or recommendations expressed in this material are those of the author(s) and do not necessarily reflect the views of the National Science Foundation.}

\bibliography{main}
\newpage
\appendix
\section{Follow the Perturbed Leader with Memory}\label{fplm:sec}
The core of our algorithm is the \citep{suggala2019online} FPL algorithm applied in the memory
setting. Here we let $\mO_{\epsilon, \mZ}$ be an optimization oracle over a convex set $\mZ$.

\begin{definition}\label{approxopt_oracle:def}
Let $z^{a} = \mO_{\epsilon, \mZ} (f)$ be the output of the approximate optimization oracle. 
Then, we have $z^{a} \in \mZ$, and 
\begin{align*}
    \sup_{z \in \mZ} f(z) - f(z^{a}) \leq \epsilon
\end{align*}
\end{definition}

\begin{algorithm}[h]
\caption{FPL with memory}
\label{FPL:alg}
\begin{algorithmic}
\STATE \textbf{Input:} Rounds $T$, history length $H$, dist. parameter $\eta$, approximate opt oracle $\mO_{\epsilon}$
\STATE Initialize $z_0, \dots ,z_{H-1}$ randomly 
\FOR {$t=H$ \TO $T$}
\STATE Play $z_t$ and suffer loss $f(z_{t-H}, \dots z_t)$
\STATE Define $g_t(z) = f_t(z, \dots z)$
\STATE Generate random vector $\sigma_t$ such that $\sigma_{t,i} \sim \Exp(\eta)$
\STATE Update $z_{t+1} \gets \mO_{\epsilon, \mZ}(\sum_{s=1}^t g_t - \inp{\sigma_t}{\cdot})$
\ENDFOR
\end{algorithmic}
\end{algorithm}

\begin{theorem}
Let $\mZ \subseteq \RR^d$ be a convex decision set with $\ell_{\infty}$ diameter at most $D$. 
Suppose the losses are $L$-Lipschitz wrt. the $\ell_1$ norm. 
For any $\eta$, Alg.~\ref{FPL:alg} achieves expected regret
\begin{align*}
    \E[\sum_{t=H}^H f_t(z_{t-H}, \dots z_t)] - \min_{z \in \mZ} \sum_{t=H}^H f(z, \dots ,z) \leq L^2H^3d^2D\eta T  + \frac{dD}{\eta} +\epsilon HT
\end{align*}
\end{theorem}

\begin{proof}
We first note that Alg.~\ref{FPL:alg} is the same approximate FPL algorithm as Alg.~1 from
\cite{suggala2019online} with loss functions $g_t$. We note that $g_t$ is $LH$-Lipschitz wrt. the
$\ell_1$ norm as 
\begin{align*}
    |g_t(x) - g_t(y)| = |f_t(x, \dots x) - f_t(y, \dots y)| \leq L \|(x,\dots x) -(y, \dots y)\|_1 = LH \|x-y\|_1~.
\end{align*}
Now, by Lemma~4 from \cite{suggala2019online}, we have that
\begin{align}\label{g_regret:eq}
    \E[\sum_{t=H}^H g_t(z_t)] - \min_{z \in \mZ} \sum_{t=H}^H g_t(z) \leq LH \sum_{t=H}^T \E[\|z_t - z_{t+1}\|_1] + \frac{dD}{\eta} +\epsilon T~.
\end{align}

To get a regret bound for $f_t$, we need to bound the difference between $f_t(z_{t-H}, \dots z_t)$
and $g_t(z_t)$. 
\begin{align*}
    |\sum_{t=H}^T f_t(z_{t-H}, \dots z_t) - \sum_{t=H}^Tf_t(z_t \dots z_t)| &\leq L \sum_{t=H}^T\sum_{j=1}^H \|z_t-z_{t-j}\|_1 \\
    &\leq L \sum_{t=H}^T \sum_{j=1}^H\sum_{l=1}^j \|z_{t-l+1} - z_{t-l}\|_1 \\
    &\leq LH^2 \sum_{t=H}^T \|z_t - z_{t+1}\|_1
\end{align*}
Now adding this to \eqref{g_regret:eq}, yields

\begin{align}\label{f_regret:eq}
    \E[\sum_{t=H}^H f_t(z_{t-H}, \dots z_t)] - \min_{z \in \mZ} \sum_{t=H}^H f(z, \dots ,z) \leq (LH^2+LH) \sum_{t=H}^T \E[\|z_t - z_{t+1}\|_1] + \frac{dD}{\eta} +\epsilon T
\end{align}

We can then follow Thm.~1 of \cite{suggala2019online} to bound 

\begin{align*}
    \E[\|z_t - z_{t+1}\|_1] \leq 125\eta LH d^2D + \frac{\epsilon}{20LH}
\end{align*}

Substituting into \eqref{f_regret:eq}, yields the result.
\end{proof}
\begin{corollary}\label{FTPL:cor}
Let $\mZ \subseteq \RR^d$ be a convex decision set with $\ell_{\infty}$ diameter at most $D$. 
Suppose the losses are $L$-Lipschitz wrt. the $\ell_1$ norm. 
With $\eta = \Theta(L^{-1}H^{-3/2}d^{-1/2}T^{-1/2})$ and $\epsilon= \Theta(T^{-1/2})$,
Alg.~\ref{FPL:alg} achieves expected regret
\begin{align*}
    \E[\sum_{t=H}^H f_t(z_{t-H}, \dots z_t)] - \min_{z \in \mZ} \sum_{t=H}^H f(z, \dots ,z) \leq O(LDd^{3/2}H^{3/2}\sqrt{T})
\end{align*}
\end{corollary}
\section{OTR with memory algorithm}\label{otrm_alg:sec}
The trust region with memory problem can be solved by applying Alg.~\ref{FPL:alg} with a fast
approximate trust region optimization algorithm in place of the optimization oracle. The $\treg$
algorithm from \cite{Hazan_2015} can be used here choosing a sufficiently low failure probability.

Alg.~\ref{OTR:alg} is a concrete instance of Alg.~\ref{FPL:alg} using a trust region solver. We start with the expansion of $g_t$ from Alg.~\ref{FPL:alg}. 

\begin{align}
    g_t(z) &= f_t(z,\dots z) \nonumber\\
    &= (z,\dots z)^{\top} P_t (z,\dots z) + p_t^{\top} (z,\dots z) \nonumber\\
    &= \begin{bmatrix}
z\\
z\\
\vdots\\
z
\end{bmatrix}^{\top}\begin{bmatrix}
P^{11}_t & P^{12}_t &\dots &P^{1H}_t\\
P^{21}_t &P^{22}_t &\dots &P^{1H}_t\\
\vdots &\vdots &\ddots &\vdots\\
P^{H1}_t &P^{H2}_t &\dots &P^{HH}_t
\end{bmatrix}\begin{bmatrix}
z\\
z\\
\vdots\\
z
\end{bmatrix} + \begin{bmatrix}
p^{1}_t \\
p^{2}_t \\
\vdots \\
p^{H}_t 
\end{bmatrix}^{\top}\begin{bmatrix}
z\\
z\\
\vdots\\
z
\end{bmatrix} \nonumber\\
&= z^{\top} \bigg ( \sum_{i=1}^{H}\sum_{j=1}^{H} P^{ij}_{t}\bigg ) z +
z^{\top}\bigg (\sum_{i=1}^{H} p^{i}_{t} \bigg) \nonumber \\
&\eqdef z^{\top} C_t z + d_t^{\top}z \label{trust_reg_g_advanced:eq}
\end{align}

\begin{algorithm}[h]
\caption{Online Trust Region}
\label{OTR:alg}
\begin{algorithmic}
\STATE \textbf{Input:} Rounds $T$, history length $H$, noise parameter $\eta$,  error tolerance $\epsilon$, $\ell_2$ bound $D$ \\
\STATE Initialize $z_0, \dots ,z_{H-1}$ randomly with $\|z_i\|\leq D$ 
\STATE $S_{H-1} = 0_{d \times d}, s_{H-1} =0_{d}$
\FOR {$t=H$ \TO $T$}
\STATE Play $z_t$ 
\STATE Observe $P_t \in \RR^{dH \times dH}$, $p_t\in \RR^{dH}$
\STATE Suffer loss $f_t(z_{t:H}) = z_{t:H}^{\top} P_t z_{t:H} + p_t^{\top} z_{t:H}$
\STATE Define $C_t = \sum_{i=1}^{H}\sum_{j=1}^{H} P^{ij}_{t}$ and $d_t = \sum_{i=1}^H p^{i}_{t}$ \quad [See \eqref{trust_reg_g_advanced:eq}]
\STATE Define $S_t = S_{t-1} + C_t$ and $s_t = s_{t-1} + d_t$
\STATE Generate random vector $\sigma_t \in \RR^d$ such that $\sigma_{t,i} \sim \Exp(\eta)$
\STATE Update $z_{t+1} \gets \treg(S_t, s_t - \sigma_t, D, \epsilon)$
\ENDFOR
\end{algorithmic}
\end{algorithm}

\begin{theorem}\label{otrm:theorem}
Suppose elements of matrices $P_t$ and elements of $p_t$ bounded by $R$. Let 
$\eta = \frac{R }{d^{3/2} DH^{3/2}} \cdot \frac{1}{ \sqrt{T}} \ \ \ \  \Theta(R(dD + 1)^{-1}H^{-3/2}d^{-1/2}T^{-1/2})$, $\eps = \Theta(T^{-1/2})$. Alg.~\ref{OTR:alg} will incur expected regret at most
\begin{align*}
    \regret(T) \leq O(R D^2 d^{5/2}H^{3/2}\sqrt{T})~,
\end{align*}
and the runtime of the algorithm for each iteration will be $\poly(d,H, \log R,\log D, \log T)$.
\end{theorem}
\begin{proof}
First note that Alg.~\ref{OTR:alg} is a correct implementation of FTPL with
$f_t$ and $g_t$ as defined in \eqref{quad_loss:eq} and \eqref{trust_reg_g_advanced:eq}. In particular,
$\treg(S_t, s_t - \sigma_t,D, \eps)$ approximately solves
\begin{align*}
    &\max_{\|z\| \leq D} z^{\top} S_t x + s^{\top}_t z - \sigma_t^{\top} z \\
    = &\max_{\|z\| \leq D} z^{\top} \bigg(\sum_{s=H}^t C_s \bigg) z + z^{\top} \bigg(\sum_{s=H}^t d_s \bigg) - \sigma_t^{\top} x\\
    = &\max_{\|z\| \leq D} \sum_{s=H}^t (z^{\top}C_s z + d_s^{\top}z) - \sigma_t^{\top} z \\
    = & \max_{\|z\| \leq D} \sum_{s=H}^t g_s(z) - \sigma_t^{\top} z~.
\end{align*}

We first note that the $\ell_{\infty}$ diameter of the constraint set, $D_\infty$ is $2D$. The next step is to bound the $\ell_1$, Lipschitz constant of $f_t(x)$. We note that $\nabla f_t(x) = (P + P^{\top})x + p$ so 
\begin{align}
     L &= \max_x \|\nabla f_t(z)\|_{\infty} \nonumber \\
     &\leq \max_x (\|P\|_{\infty}+ \|P\|_{1}) \|z\|_{\infty} + \|p\|_{\infty} \nonumber\\
     &\leq (\|P\|_{1} + \|P\|_{\infty})D + \|p\|_{\infty}\nonumber\\
     &\leq 2dRD + R \label{l1_lipschitz:eq}
\end{align}

The regret bound follows after applying Cor.~\ref{FTPL:cor} and \eqref{l1_lipschitz:eq}. To bound the runtime of an iteration, we need to find the condition number for the trust region instances passed to $\treg$. Because the $S$ matrices sum over time, 
it can be seen that the condition number can grow at most linearly in $T$. The runtime then follows from Thm.~\ref{tregion_opt:theorem}.
\end{proof}
\section{Approximating the state and costs with a truncated rollout}\label{approx_state:sec}
We return to the control application. We want to analyze the state $x^{\A}_{t+1}$ that results from a noise generator $\A$ that plays CDG $M_t = (M^{[1]}_t \dots M^{[H]}_t)$. Unrolling the LDS recursion $H$ times can be done to yield the following transfer matrix $\Psi_{t,i}$, describing the effect of $u_{t-i}$ on the state $x_{t+1}$. 

\begin{definition}\label{transfer_func:def}
Define the disturbance-noise transfer matrix $\Psi_{t,i}$ to be
\begin{align}\label{transfer_func:eq}
    \Psi_{t,i}(M_{t-H} \dots M_{t-1}) = A^{i}B\mathbf{1}_{i \leq H} + \sum_{j=1}^H A^{j-1}M_{t-j}^{[i-j]} \mathbf{1}_{i-j \in [1,H]}
\end{align}
\end{definition}

We note that $\Psi_{t,i}$ is linear in $(M_{t-H} \dots M_{t-1})$. 

\begin{lemma}\label{state_ev:lem}
If $w_t$ is chosen using the non-stationary noise policy $(\rho_0 \dots \rho_{T-1})$, then the state sequence satisfies
\begin{align}\label{trunc_transfer_expansion:eq}
    x^{\A}_{t+1} = A^{H+1}x^{\A}_{t-H} + \sum_{i=0}^{2H} \Psi_{t,i} u_{t-i}
\end{align}
\end{lemma}

As the notion of regret is counter-factual, decisions from all time steps effect the current state and thus the current loss. To mitigate this effect, we approximate the state and cost using only the effect of the last $H$ time steps. Lem.~\ref{approx_cost:lem} shows that such an approximation is valid, as the current state has a negligible dependence on the long term past. 

\begin{definition}\label{approx_state:def}
Define an approximate state $y_{t}$, which is the state the system would have reached if it played the non-stationary policy $(M_{t-H-1} \dots M_{t-1})$ at all time steps from $t-H$ to $t$ assuming $y_{t-H}=0$. we have

\begin{align*}
    y_{t}(M_{t-H-1:t-1}) = \sum_{i=0}^{2H} \Psi_{t-1,i} u_{t-i-1}~,
\end{align*}
where we use the notation $M_{t-H-1:t-1} = [M_{t-H-1} \dots M_{t-1}]$. The approximate cost is then given by
\begin{align*}
    f_{t}(M_{t-H:t}) = c_{t}(y_{t}(M_{t-H:t}), u_t) = \|y_{t}(M_{t-H-1:t-1})\|^2_{Q_t} + \|u_t\|^2_{R_t}~.
\end{align*}

\end{definition}

\begin{lemma}\label{approx_cost:lem}
Suppose $\A$ plays according to $(M_{t-H-1} \dots M_{t-1})$ at all time steps from $t-H$ to $t$, Assumptions~\ref{bounded-controls:as}, \ref{stab_dy:as}, and \ref{bounded_costs:as} holds and $H = \ceil{\gamma^{-1}\log(\kappa \xi T)}$, then we have that $|c_t(x^{\A}_t) - f_t(M_{t-H:t})| \leq \frac{C_x}{T}$, where $C_x = \frac{2\beta H D C_u}{\gamma}$.
\end{lemma}

\begin{proof}
We first show that the states are bounded.  We note that $w_t = M_t u_{t-H:t-1}$ so $\|w_t\| \le H D C_u$ by applying Assumption~\ref{bounded-controls:as} and the Definition~\ref{comparator:def}.  Now applying Assumption~\ref{stab_dy:as} and triangle inequality, we have $\|Bu_t + Cw_t\| \le 2\beta H D C_u$.  The state $\|x^{\A}_{t-H-1}\| \le \frac{2\beta H D C_u}{\gamma}= C_x$ via strong stability of $A$.
We first note that 
\begin{align*}
    \|x^{\A}_t - y^{M_{t-H:t}}_t\| &= \|A^{H+1}x^{\A}_{t-H-1}\| &[\text{Lem.}~\ref{state_ev:lem}]\\
    &\leq \|A^{H+1}\|\|x^{\A}_{t-H-1}\|\leq  \|A^{H+1}\|C_x \\
    &\leq \|H\|\|H^{-1}\|\|L\|^{H+1}\|x^{\A}_{t-H-1}\|\leq \kappa(1-\gamma)^{H+1} C_x &[\text{Assumption}~\ref{stab_dy:as}]\\
    &\leq \kappa C_x e^{-\gamma H} &[1+x \leq e^x]
\end{align*} 

Now, we use Assumption~\ref{bounded_costs:as} to complete the result.
\begin{align*}
    |c_t(x^{\A}_t) - f_t(M_{t-H:t})| = |c_t(x^{\A}_t) - c_t(y_t))| \leq \xi \|x^{\A}_t - y_t\|^2 \leq \xi \kappa C_x e^{-\gamma H} \leq \frac{C_x}{T}
\end{align*}

\end{proof}

\section{Applying Alg.~\ref{OTR:alg} to approximate costs}\label{main_proof:sec}
We apply Alg.~\ref{OTR:alg} with the approximate cost $f_t$ as defined in Def.~\ref{approx_state:def}. To do this, we derive a closed form expression for $g_t(M)= f_t(M, \dots M)$. The transfer matrix can be simplified to $\Phi_{t,i}(M\dots M) = C_iM + D_i$, where $C_i\in \RR^{d_x \times Hd_x}$ is a block matrix with blocks either $0$ or powers of $A$, and $D_i= A^{i}B\mathbf{1}_{i \leq H}$. We can write $g_t$ as a quadratic function of $M$ as follows
\begin{align}
    g_t(M) &= \|y_{t}(M \dots M)\|^2_{Q_t} + \|u_t\|^2_{R_t}  \nonumber\\
           &=  \|\sum_{i=0}^{2H} C_i M u_{t-i-1} + D_i u_{t-i-1}\|^2_{Q_t}  + \|u_t\|^2_{R_t} \nonumber\\
           &= \vect(M)^{\top} P_t \vect(M) + p_t^{\top} \vect(M) + c~. \label{gtcontrol:eq}
\end{align}

Here $P_t \in \RR^{Hd_xd_u \times Hd_xd_u}$ and $p_t \in \RR^{Hd_xd_u}$ are defined by
\begin{align}
    P_{t, k + Hd_x(l-1), m + Hd_x(n-1)}  &= \sum_{i=0}^{2H}\sum_{j=0}^{2H} \pa{U^{t}_{ij}}_{ln}\pa{C^{\top}_i Q_t C_j}_{km} \label{At:eq}\\
    p_t &=  2\sum_{i=0}^{2H}\sum_{j=0}^{2H}\vect(C^{\top}_iQ^{\top}_tD_j U^{t}_{ij})\label{bt:eq}~,
\end{align}
where $U^t_{ij} = u_{t-j-1} u_{t-i-1}^{\top}$.

Applying the closed form for $g_t$ in Alg.~\ref{OTR:alg} yields Alg.~\ref{ODG:alg}.

\begin{lemma}\label{coefficient_bound:lem}
Suppose Assumptions~\ref{bounded-controls:as} and ~\ref{bounded_costs:as} hold, then 
$f_t(M_{t-H:t})$ is the quadratic function 
\begin{align*}
    \vect(M_{t-H:t})^{\top} P_t \vect(M_{t-H:t}) + p_t^{\top}\vect(M_{t-H:t})~,
\end{align*}
 with coefficients of $P_t$ bounded by $\xi C_u^2 \kappa^2$ and coefficients of $p_t$ bounded by $\xi C_u^2 \kappa^2 \beta$.
\end{lemma}
\begin{proof}
We bound the coefficients of the quadratic functions $f_t(M_{t-H}, \dots M_t)$ by noting that in $y_t(M_{t-H}, \dots M_t)$ terms of the form $M_{t-j}^{[i-j]} u_{t-i}$ occur exactly once. Using Assumption~\ref{stab_dy:as}, we see that $y_t$ is $\kappa C_u$ Lipshitz in $M_{t-j}^{[i-j]}$. We can then bound $y_t^{\top} Q_t y_t \leq \xi \|y_t\|^{2}$ using Assumption~\ref{bounded_costs:as}, resulting in the coefficients of $A_t$ being upper bounded by $\xi C_u^2 \kappa^2$. Likewise, using the bound $\|B\| \leq \beta$, we see the coefficients of $b_t$ are bounded by $\xi C_u^2 \kappa^2 \beta$
\end{proof}
\vspace{-0.5cm}
\begin{theorem}
Suppose Assumptions~\ref{bounded-controls:as}, \ref{stab_dy:as}, and \ref{bounded_costs:as} hold, then Alg.~\ref{ODG:alg} suffers regret at most $\tO(\poly(\mL)\sqrt{T})$.
\end{theorem}
\begin{proof}
The regret bound follows via the approximation guarantee from Lem.~\ref{approx_cost:lem} along with applying Thm.~\ref{otrm:theorem} for the Online Trust Region algorithm, with bounds on the coefficients for the control setting from Lem.~\ref{coefficient_bound:lem}.
\begin{align*}
\regret_t(\A) &= \max_{M \in \Pi} \sum_{t=1}^T c_t(x^{M}_t, u_t) - \sum_{t=1}^T c_t(x^{\A}_t, u_t)& \\
    &\leq \max_{M \in \Pi} \sum_{t=1}^T (f_t(M, \dots, M) + \frac{C_x}{T}) - \sum_{t=1}^T (f_t(M_{t-H:t}) + \frac{C_x}{T}) & \mbox{[Lem.~\ref{approx_cost:lem}]}\\
    &\leq \tO(\poly(\mL)\sqrt{T}) & \mbox{[Lem.~\ref{coefficient_bound:lem} and Thm.~\ref{otrm:theorem}]}
\end{align*}
\vspace{-0.5cm}
\end{proof}
\section{Notes on Implementation} 
\subsection{Dynamics Transformation}
Our implementation utilizes Assumption \ref{stab_dy:as} to recast the dynamics and disturbance generator. In particular, for any stabilizable system, we apply a transformation to the dynamics using the $H_\infty$ optimal controller as $K^*$. Then we apply the learning algorithm over the residuals, which will be valid so long as the true controller is stabilizing. 

In particular, define $(K_H, W_H)$ as the solution to the $H_\infty$ game acting on dynamics $(A, B, C)$ for arbitrary cost matrices $Q, R$. This solution exists iff the dynamics are stabilizable. Then we apply the following transformation: 
\begin{equation*}
    \begin{split}
    x_{t+1} & = Ax_t + Bu_t + Cw_t \\
    & = (A-BK_H)x_t + B(K_Hx_t + u_t) + Cw_t \\
    & = \tilde{A}x_t + Bu'_t + Cw_t
    \end{split}
\end{equation*}
Here, $\tilde{A}$ is stable and the main analysis can proceed. 

Ideally, our parameterization of the disturbances would be guaranteed to recover the $H_\infty$ Nash equilibrium against an $H_\infty$ controller. Therefore, we let the bias term of our algorithm be $W_Hx_t$. Defining $r_t = u'_t = K_Hx_t + u_t$, we see that for $u_t = -K_Hx_t$, we have $r_t = 0$ and recover the desired behavior: 
\begin{equation*}
\begin{split}
    x_{t+1} & = \tilde{A}x_t + Br_t + Cw_t \\
    & = \tilde{A}x_t + C(W_Hx_t + \sum_i M^{[i]} r_{t-i}) \\
    & = \tilde{A}x_t + CW_Hx_t
\end{split}
\end{equation*}
For cases in which the controller deviates from the $H_\infty$ equilibrium, the $H_\infty$ generator is no longer necessarily Nash optimal, and therefore our generator will be able to deviate in order to exploit any arising weaknesses in the controller robustness. \\

Of course, it is possible to have the bias term include both an $H_\infty$ and learned component, if it is deemed wise to do so. In general, we try to avoid hyperparameter tuning of this nature. 

\subsection{Computational Notes} \label{appComp:section}
In practice the disturbances will need to be bounded, in order to prevent either unbounded growth in the disturbances, or rapid decay (depending on whether the domain is too large or too small). Setting the parameters to be small generally appears to be the better option. 

The algorithm runtime is generally short. The AirSim simulation runs in approximately real time on a CPU with a 3.30 GHz Intel i9-9820X processor. For the linear dynamical systems simulations, the four-state, 2-input case runtime is approximately 15 minutes for the simulations comprising 18 controller-generator pairs over 11 systems and 10 seeds. 

The AirSim simulation utilizes numerical integration with a timestep $dt = 0.001$ seconds. In order to analyze our system, the observed rotor speed inputs that are seen by the disturbance generators are averaged over 0.25 second intervals, with 25 data points per interval. The wind policy itself is thus only allowed to update every 0.25 seconds. This limits the wind to oscillations of at most 2 Hz, in line with observed power spectra \citep{thomas_id_96}, which show the presence of pressure fluctuations greater than 1 Hz but general attenuation above 5 Hz. 

The key hyperparameter of MOTR (with our bias implementation) is the allowable cumulative norm magnitudes of the matrices $M$ of our learned parameterization [termed $D_M$]. Because the disturbances are norm-limited, choice of $D_M$ affects the relative weighting of the trust region solver (grows with higher $D_M$) and the bias term. In general, a smaller $D_M$ prevents significant projection distances, which keeps the estimate closer to that of the regret minimizing theoretical generator and improves performance. In many cases, even small $D_M$ allow significant differentiation between MOTR and the $H_\infty$ generator, which is the limiting case as $D_M \rightarrow 0^+$. 

\end{document}